\documentclass{article} 
\usepackage{iclr2020_conference,times}
\usepackage{CJKutf8}
\usepackage{hyperref}

\hypersetup{
    colorlinks,
    citecolor=blue,
    filecolor=black,
    linkcolor=black,
    urlcolor=black,
    linktoc=all
}

\usepackage{bookmark}
\usepackage{url}
\usepackage{graphicx}
\usepackage{amsfonts}
\usepackage{amsmath}
\usepackage{booktabs}
\usepackage{listings}
\usepackage{amsmath}
\usepackage{bbm}
\usepackage{xcolor}
\usepackage{enumitem}
\usepackage{booktabs}
\usepackage[most]{tcolorbox}
\usepackage{color, colortbl}
\usepackage{float}
\usepackage{multirow}
\usepackage{listings}
\usepackage{makecell}
\usepackage{enumitem}
\usepackage{pifont}

\UseRawInputEncoding

\definecolor{codegreen}{rgb}{0,0.5,0}
\definecolor{codeblue}{rgb}{0.25,0.5,0.5}
\definecolor{codegray}{rgb}{0.6,0.6,0.6}

\usepackage[titles]{tocloft}
\title{A Review of Sparse Expert Models in \linebreak Deep Learning}

\author{
William Fedus\thanks{Equal contribution. Correspondence to \{\texttt{liam.fedus,barretzoph\}@gmail.com}.} \\
Google Brain \\
\And
Jeff Dean \\
Google Research \\
\And
Barret Zoph\footnotemark[1] \\
Google Brain \\
}

\iclrfinalcopy 

\begin{document}
\maketitle

\begin{abstract}
Sparse expert models are a thirty-year old concept re-emerging as a popular architecture in deep learning.
This class of architecture encompasses Mixture-of-Experts, Switch Transformers, Routing Networks, BASE layers, and others, all with the unifying idea that each example is acted on by a \emph{subset} of the parameters.
By doing so, the degree of sparsity decouples the parameter count from the compute per example allowing for extremely large, but efficient models.
The resulting models have demonstrated significant improvements 
across diverse domains such as natural language processing, computer vision, and speech recognition.
We review the concept of sparse expert models, provide a basic description of the common algorithms, contextualize the advances in the deep learning era, and conclude by highlighting areas for future work.
\end{abstract}

\section{Introduction}

Remarkable advances in machine learning -- especially in natural language -- have been achieved by increasing the computational budget, training data, and model size. 
Notable milestone language models include GPT-2 \citep{radford2018improving},
BERT \citep{devlin2018bert},
T5 \citep{raffel2019exploring}, 
GPT-3 \citep{brown2020language}, 
Gopher \citep{rae2021scaling},
Chinchilla \citep{hoffmann2022training},
and PaLM \citep{chowdhery2022palm}.
However, state-of-the-art models now require thousands of specialized, interconnected accelerators for weeks or months at a time. 
These models are therefore expensive to produce and incur high energy costs \citep{patterson2021carbon}. 
Therefore, as the scale of machine learning systems has increased, the field has sought more efficient training and serving paradigms.
Sparse expert models have risen as a promising solution.
\begin{figure}[ht!]
    \centering
    \includegraphics[width=\textwidth]{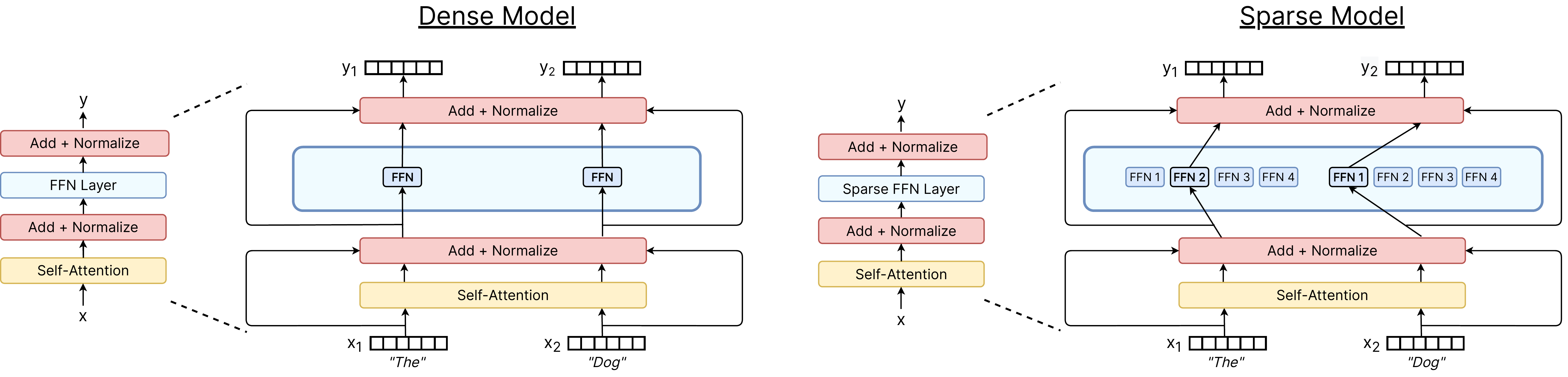}
    \caption{\textbf{Comparing a dense and sparse expert Transformer.} A dense model \textbf{(left)} sends both input tokens to the same feed-forward network parameters (FFN). A sparse expert model \textbf{(right)} routes each input token independently among its four experts (FFN1 $\cdots$ FFN4). In this diagram, each model uses a similar amount of computation, but the sparse model has more unique parameters.
    Note while this figure showcases a specific and common approach of sparse feed-forward network layers in a Transformer \citep{vaswani2017attention}, the technique is more general.}
    \label{fig:sparsity_diagram}
\end{figure}

Sparse expert models, of which, Mixture-of-Experts (MoE) is the most popular variant, are neural networks where a set of the parameters are partitioned into ``experts'', each with a unique weight.
During training and inference, the models route input examples to specific expert(s) weights.
As a result, each example only interacts with a \emph{subset} of the network parameters, contrasting the usual approach where the entire network is used for each input.
Because only a fraction of the experts are used for each example, the amount of computation may remain small relative to the total model size.

Many modern sparse expert models draw inspiration from \cite{shazeer2017outrageously}, which trained the largest model at the time and achieved state-of-the-art language modeling and translation results.
Sparse expert models have further surged in popularity when combined with Transformer language models \citep{lepikhin2020gshard,fedus2021switch}.
And while most work has been in natural language processing, they have also been successfully used in a variety of domains including computer vision \citep{puigcerver2020scalable}, speech recognition \citep{you2021speechmoe} and multi-modal learning \citep{mustafa2022multimodal}.
Recent work by \cite{clark2022unified} rigorously studied the scaling properties of sparse expert models across different model sizes and number of experts. 
Further, state-of-the-art results on many benchmarks are currently held by sparse expert models such as ST-MoE \citep{zoph2022designing}.
The field is evolving quickly with research and engineering advances increasing our understanding and improving empirical results.

We narrow our survey to sparse expert models in the era of deep learning (heuristically 2012-onward), recounting recent advances and discussing promising future avenues.
For a comprehensive review of the history of Mixture-of-Experts, predating the recent deep learning advances, we refer readers to the survey, ``Twenty Years of Mixture-of-Experts'' \citep{yuksel2012twenty}.
Further, sparse expert models may be regarded as a special class of adaptive computation models which are surveyed in \cite{xu2022survey}.
Finally, \cite{tay2020efficient} surveys a broader set of methods aimed at increasing the computational efficiency of Transformers, of which, sparse expert models are one promising approach.

\section{Sparse Expert Models} 
The concept of MoE in machine learning dates back at least three decades to the work of \cite{jacobs1991adaptive,jordan1994hierarchical}.
In early concepts, the experts defined an entire neural network and the MoE was similar to ensemble methods.

\subsection{In Deep Learning}
\cite{eigen2013learning} proposed architectures that used stacked layers of Mixture-of-Experts on jittered MNIST \citep{lecun1998gradient}.
This work used a continuous mixture of the experts'’ outputs (soft selection) rather than restricting to the top subset of experts at each layer (hard selection) -- limiting its practicality\footnote{The full computational cost is incurred with soft selection even if the expert was not necessary (i.e. an exceedingly small routing weight).}. 
This work, however, set the stage for later efficient implementations which relied on the idea of MoE as a \emph{component} of a neural network.
The first large-scale success of this approach in deep learning came from \cite{shazeer2017outrageously}.
This work inserted an MoE layer between two LSTM layers \citep{hochreiter1997long} where the output from the lower layer's LSTM was sent for computation within the MoE.
The resulting sparse model was state-of-the-art in machine translation, though the largest variant with 131,072 experts and 137B parameters generalized worse than smaller variants.
Despite this success, however, follow-on research was relatively dormant with greater emphasis on directly studying the Transformer \citep{vaswani2017attention}. 
This changed with the release of GShard \citep{lepikhin2020gshard} and Switch Transformers \citep{fedus2021switch} -- both of which replaced the feed-forward layers in Transformers with expert layers.
However, while the experts-as-a-layer approach has become the dominant paradigm, more recent works revisit the concept of experts as fully independent models \citep{gururangan2021demix,li2022btm}.
This confers a benefit of modularity and composability; \cite{li2022btm} shows custom networks can be constructed by composing them of expert language models trained on specific domains.

\begin{figure}[ht!]
    \centering
    \includegraphics[width=0.65\textwidth]{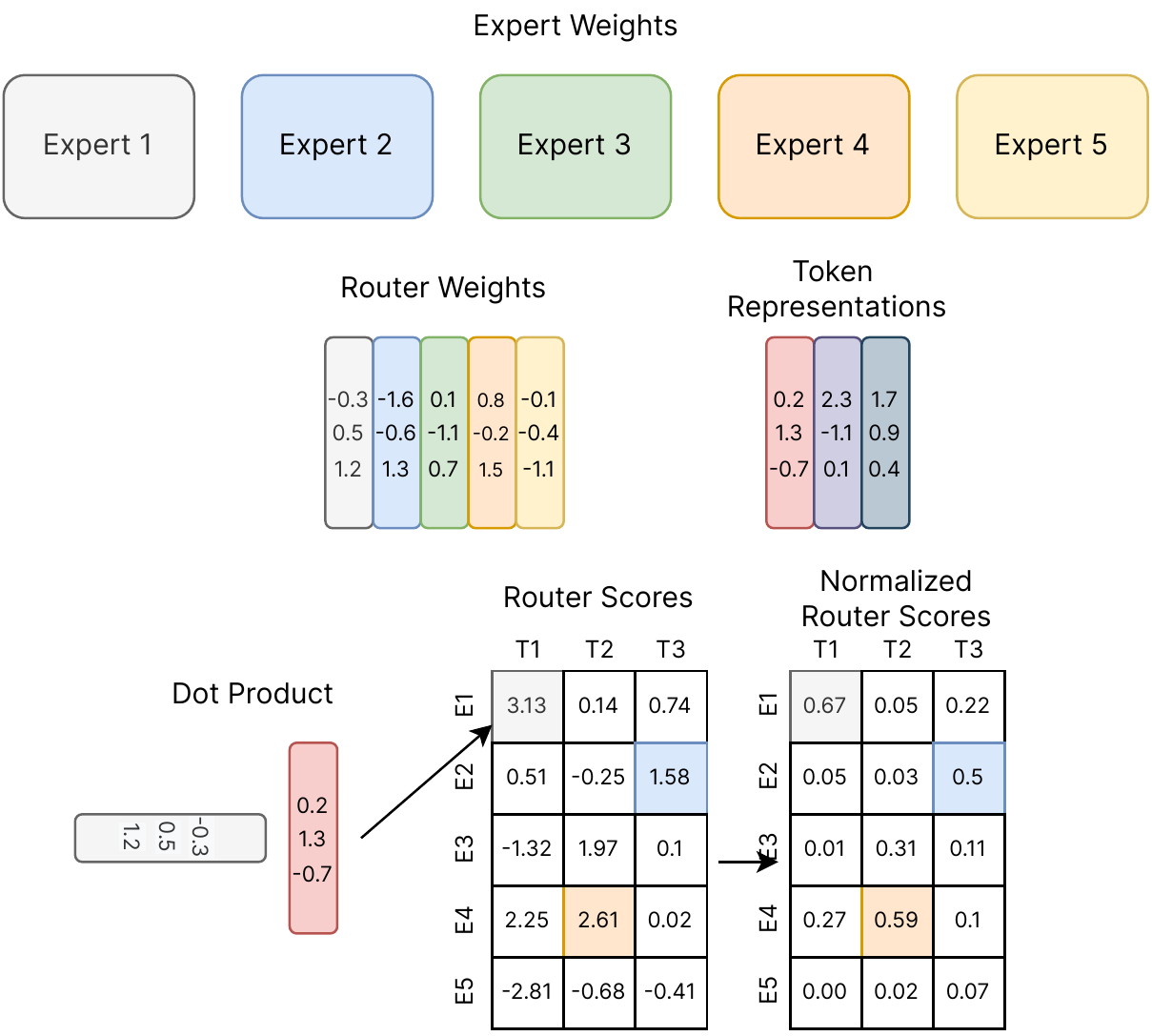}
    \caption{\textbf{Schematic of top-$k$ routing}. We visualize an example of the top-$k$ token routing scheme over five experts and three input tokens. Each expert and token is color-coded and the router weights ($W_r$) have a representation for each expert (color matched). To determine the routing, the router weight performs a dot product with each token embedding ($x$) to produce the router scores ($h(x)$). These scores are then normalized to sum to one ($p(x$)).}
    \label{fig: router_math}
\end{figure}

Figure \ref{fig: router_math} illustrates the original top-$k$ routing mechanism proposed in \cite{shazeer2017outrageously} which was foundational to many follow-on works. 
New advances to the routing algorithm are described in Section \ref{sec: routing}.
Choosing experts based on the input usually entails a discrete selection (i.e. which expert to use), which complicates backpropagation algorithms relying on differentiability.
As a solution, \citet{shazeer2017outrageously} proposed a top-$k$ routing function which takes as an input a token representation $x$ and then routes it to the top-$k$ experts out of the set $\{E_i\}_{i=1}^N$ of $N$ experts.
The router has a trainable variable $W_r$ which computes the logits  $h(x) = W_r \cdot x$ which are normalized via a softmax distribution over the $N$ experts.
The gate-value for expert $i$ is given by,
\begin{equation}
    p_i(x) = \frac{e^{h(x)_i}}{\sum_j^N e^{h(x)_j}}.
\end{equation}
We denote the set of selected top-$k$ expert indices as $\mathcal{T}$.
The output computation of the layer is the linearly weighted combination of each expert's computation on the token by the gate value,
\begin{equation}\label{eqn: moe_layer}
    y = \sum_{i \in \mathcal{T}} p_i(x) E_i(x). 
\end{equation}
We note that in contrast to \cite{eigen2013learning}, this selection is only over the top-$k$ experts and is thus more computationally efficient.

\subsection{On Modern Hardware} \label{sec: hardware}
Modern sparse expert models have been co-designed with the distributed systems used to train the largest neural networks.
These are a special case of sparse neural networks \citep{gale2019state,dettmers2019sparse,evci2020rigging} which are similar in that they only use a subset of parameters, but differ because they have potentially irregular sparsity patterns. 
And while generic sparse neural networks (with irregular sparsity patterns) reduce overall theoretical FLOPs, these are often not efficiently supported on current hardware which specialize in linear algebra operations on contiguous (regular) blocks of memory.
Sparse expert models, on the other hand, activate entire blocks of parameters (i.e.\@ entire matrices), and thus easily translate theoretical FLOPs savings to practical time savings on modern hardware \citep{fedus2021switch,rajbhandari2022deepspeed}.

The largest neural networks \citep{brown2020language,rae2021scaling,chowdhery2022palm} now far exceed the memory capacity of a single accelerator and therefore tensors (e.g. weights, activations, optimizer variables) are sharded using various parallelism strategies.
Three common approaches include data parallelism (model weights replicated, but data sharded), tensor model-parallelism \citep{shazeer2018mesh} (data and weight tensors are split across devices), and pipeline parallelism \citep{harlap2018pipedream,huang2019gpipe} (entire layers or groups of layers are split across devices).
Mixture-of-Experts fit naturally with these parallelism schemes.
Experts reside on different accelerators and the input data is dynamically dispatched to and fetched from them.
Early architectures often employed many, small experts that would fit within an individual accelerator \citep{lepikhin2020gshard}, but later works designed larger experts that must be split across accelerators \citep{fedus2021switch,du2021glam} and required additional optimizations for communication efficiency \citep{shazeer2018mesh,roberts2022t5x,rajbhandari2022deepspeed}.

Dynamic routing on distributed systems incurs additional communication overhead beyond standard Transformer models.
Dispatching the inputs to the experts is often implemented as an \texttt{all2all} communication primitive, where each accelerator communicates data to all other accelerators.\footnote{Many routing algorithms (but not all) incur two \texttt{all2all} communication costs in the forward pass and another two in the backward pass. An example of a routing algorithm using more is BASE layers \citep{lewis2021base}, which requires four \texttt{all2all} in the forward pass and another four in the backward pass.}
The \emph{capacity factor} directly impacts the communication cost by modulating the expert batch size \citep{lepikhin2020gshard} to be $CF \cdot \left(B / E \right)$, where $CF$ is the capacity factor, $B$ is the total tokens per batch and $E$ is the number of experts.
Larger capacity factor values can improve quality, but at the expense of increased communication, memory and compute costs.
Efficient implementations of the \texttt{all2all} primitive, along with changes to the routing algorithm (e.g. reduced capacity factor), alleviate the added communication costs from sparse expert algorithms.

When training normal distributed Transformers it is known in advance what batch of data each accelerator will process.
However, dynamic routing algorithms break this property because inputs are dynamically routed to experts, which can often lead to different number of inputs getting sent to each of the experts.
Therefore, routing algorithms often encourage \emph{load balance} over the accelerators to encourage good utilization.
Load balance has been accomplished by auxiliary losses \citep{shazeer2017outrageously} as well as through treating this as a linear assignment problem \citep{lewis2021base,clark2022unified}.
More details on advances to load balancing are provided in Section \ref{sec: routing}.

Finally, recent systems advances have further improved both the training and deployment of MoE models.
\cite{jaszczur2021sparse} sparsify all the layers (e.g. dense and self-attention) of a Transformer model to achieve 37$\times$ inference speedups for a special-case of single-example inference (unbatched).
\cite{kossmann2022optimizing} relaxes the constraints of static expert batch sizes with the \textsc{RECOMPILE} library.
This system dynamically recompiles and optimizes the computational resources of Mixture-of-Experts models so tensor sizes are matched to the experts' computational demands, not statically-set arrays.
Next, in addition to data-, model-, and expert-parallelism, the DeepSpeed-MoE library \citep{rajbhandari2022deepspeed} supports ZeRO partitioning \citep{rajbhandari2019zero} (fully partition tensors and regather as needed) and ZeRO-Offload (offloading to CPU to reduce GPU memory usage).
This system yielded 10$\times$ inference improvements \citep{rajbhandari2022deepspeed} and state-of-the-art translation \citep{kim2021scalable} -- increasing the practicality of these models for production services.

\section{Scaling Properties of Sparse Expert Models}
The cross-entropy loss of dense neural language models was shown to scale as a power-law (i.e. $l(x) = \left(c / x \right) ^ \alpha$ for a variable $x$) with respect to the model parameter count, amount of data, and compute budget when not constrained by the other two factors \citep{kaplan2020scaling}.
The power law coefficients were later corrected in \cite{hoffmann2022training}, which demonstrated that compute-optimal models required a closer balance of data and parameter scaling.
In contrast, early research in sparse expert models scaled heuristically -- achieving strong empirical results -- but without careful characterization of the scaling laws.
Further, several works highlighted discrepancies between upstream (e.g. pre-training) and downstream (e.g. fine-tuning) behavior \citep{fedus2021switch,artetxe2021efficient}, further complicating the understanding and explanation of sparse expert models.

\subsection{Upstream Scaling}
Sparse expert models have excelled when trained on large datasets.
A common paradigm in natural language processing is to perform upstream training (e.g. pre-training) which is then followed by downstream training (e.g. fine-tuning) on data distributions of specific interest.
Sparse expert models have consistently yielded high gains over dense counterparts during the upstream phase.
\cite{shazeer2017outrageously} presented scaling curves with respect to model parameters and the computational budget on the 1-Billion-Word Language-Modeling Benchmark \citep{chelba2013one}, achieving significant gains over dense versions.
\cite{lepikhin2020gshard} presented translation improvements as a function of model scale, and obtained a 13.5 BLEU score gain on their largest 600B parameter sparse model.
Switch Transformers \citep{fedus2021switch} measured 4-7$\times$ speed-ups in wall-time using the same compute resources over T5 models. 
The work also studied the cross entropy loss scaling as a function of parameter count, but observed the gains diminished with 256+ experts.

Furthering our understanding, \cite{artetxe2021efficient} distinguished upstream scaling behavior of MoE models on \emph{in-domain} and \emph{out-of-domain} data and found significantly better scaling for in-domain language modeling compared to dense models, corroborating the difficulties of transfer from \cite{fedus2021switch}.

\begin{figure}[ht!]
    \centering
    \includegraphics[width=0.45\textwidth]{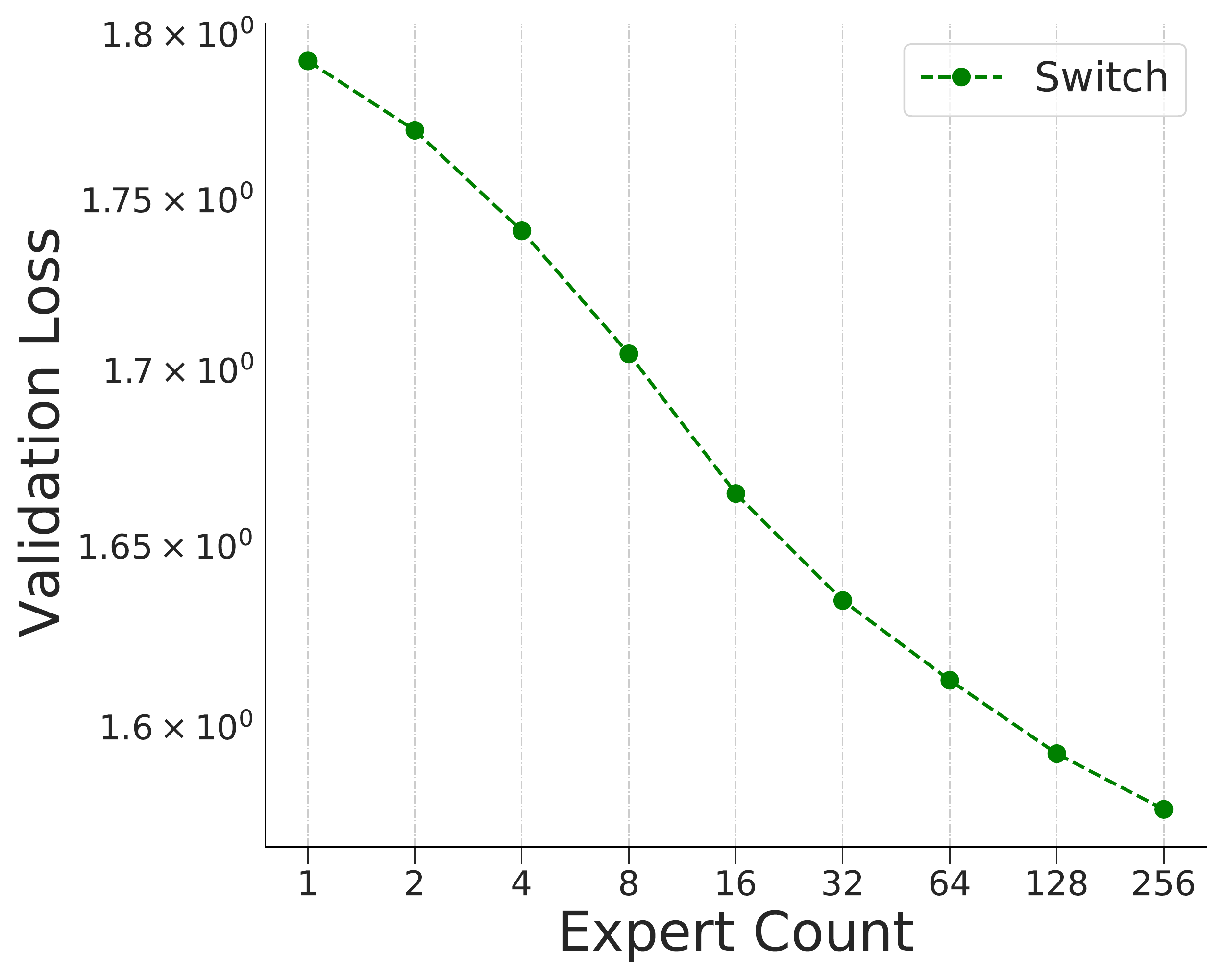}
    \includegraphics[width=0.45\textwidth]{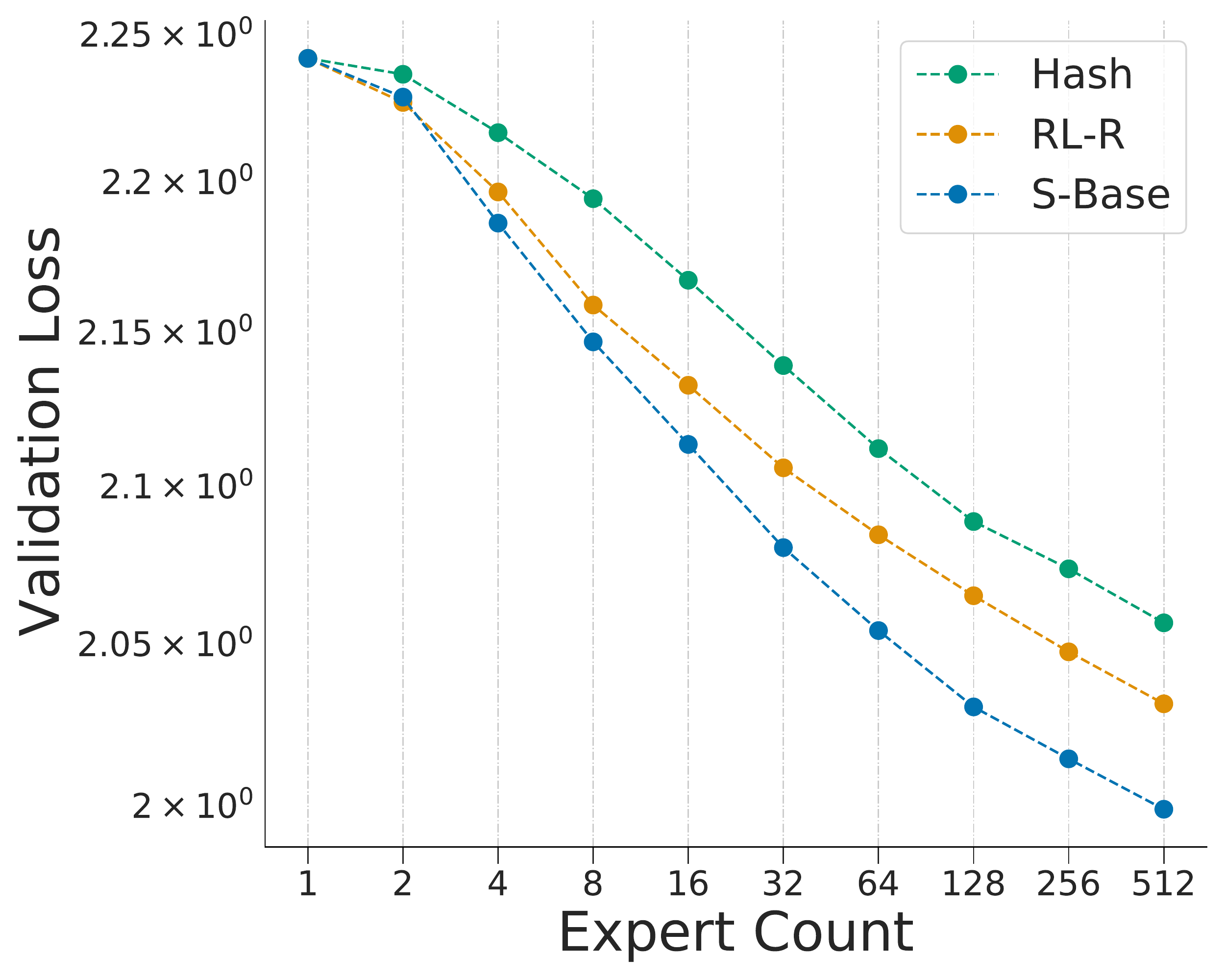}
    \caption{\textbf{Sparse scaling plots with expert count}. The cross-entropy scaling plots as a function of the number of experts are shown from \cite{fedus2021switch} \textbf{(left)} and the three sparse variants from \cite{clark2022unified}, S-Base, RL-R, Hash \textbf{(right)}. The top left-most point in both plots is an approximately compute-matched dense model. As the expert count increases, the models become increasingly sparse and yield lower validation losses.}
    \label{fig:sparse_scaling}
\end{figure}

After these early empirical successes, \cite{clark2022unified} conducted the first large-scale effort to mathematically characterize the scaling properties of sparse expert models.
This work considered three classes of sparse models and derived a notion of \emph{effective parameter count} (EPC).
The EPC estimates the dense-parameter equivalent for a sparse expert models, based on the FLOPs and the number of experts.
It was derived by conjecturing that sparse expert models followed a bilinear loss and it was shown empirically that the cross entropy loss scales as a power law in this variable.
Figure \ref{fig:sparse_scaling} presents the cross entropy scaling of Switch Transformers on the left and the three sparse variants of \cite{clark2022unified} on the right.

One key property of the scaling curves was that the gain of sparse expert models decreased with scale, which when extrapolated, implied that there would be no further benefit of sparsity beyond 900B parameters of FLOPs.
This result, however, was dependent on the number of tokens used for training and all models used only 130B tokens.
But in light of the recent scaling results from \cite{hoffmann2022training} which recommends more tokens to train compute-optimal models (Chinchilla was a 70B parameter model trained on 1.4T tokens), future work might revisit this analysis.

\subsection{Downstream Scaling}
However, the reliable upstream scaling did not immediately yield consistent gains on downstream tasks.
In one work highlighting the challenge of transfer, \cite{fedus2021switch} observed 4$\times$ pre-training improvements with a low-compute, high-parameter encoder-decoder Transformer (1.6T parameters with 2048 experts per sparse layer), but it fine-tuned poorly on reasoning-heavy tasks such as SuperGLUE \citep{wang2019superglue} compared with dense models. 
This finding hinted at further necessary research as well as a potential needed balance between computation and parameters.
However, strong empirical results soon followed in few-shot inference, fine-tuning, and other modalities.

\begin{figure}[ht!]
    \centering
    \includegraphics[width=0.45\textwidth]{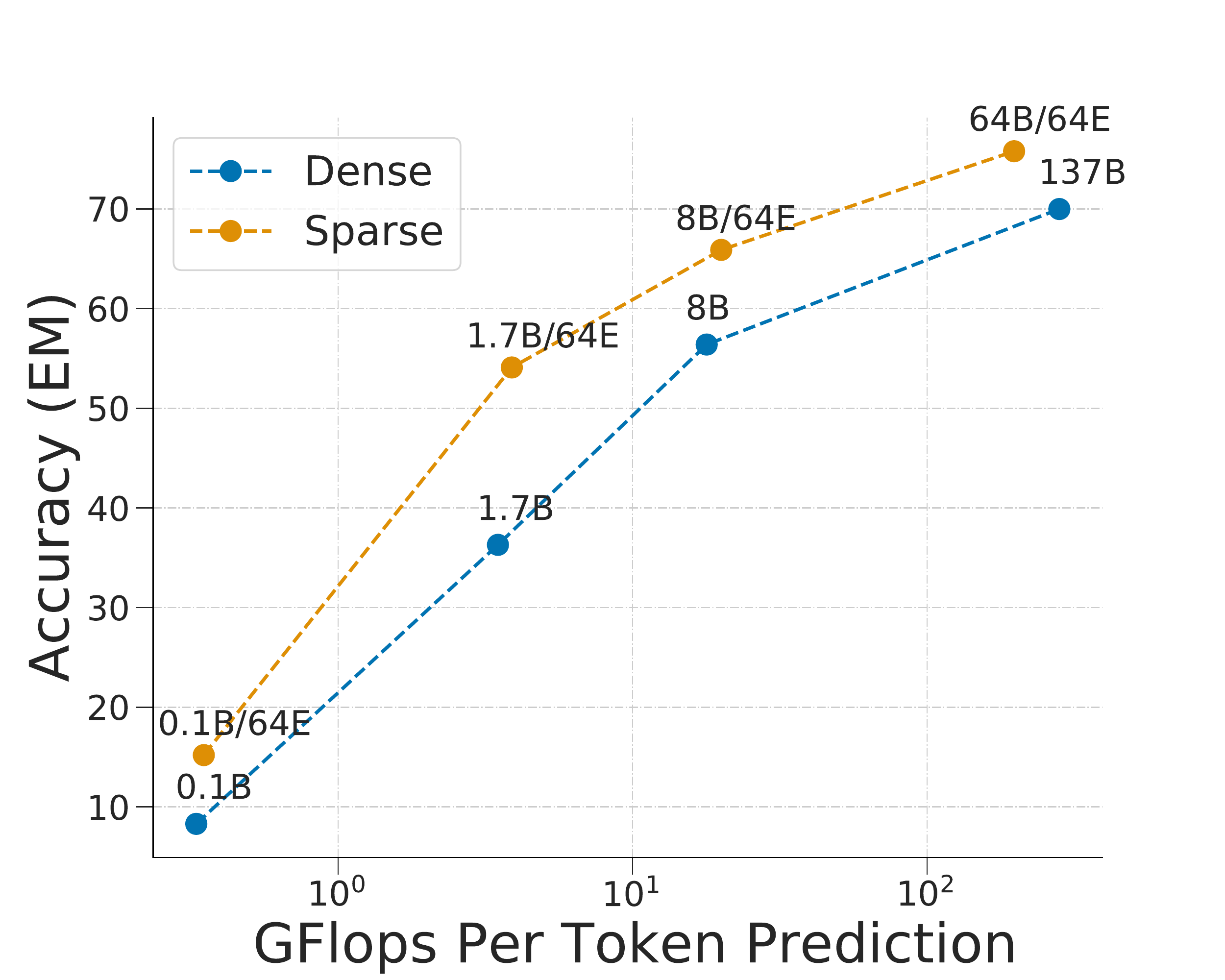}
    \includegraphics[width=0.45\textwidth]{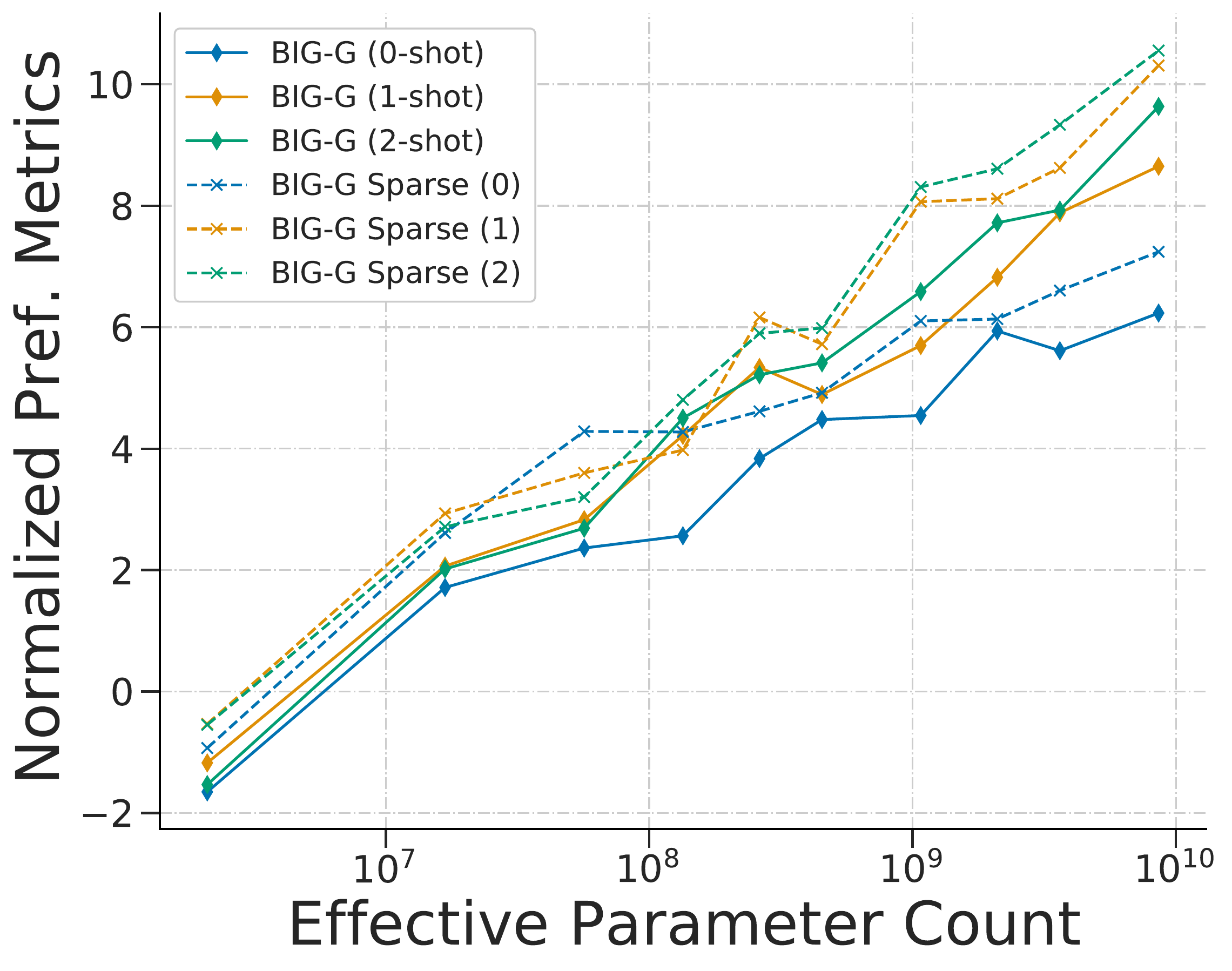}
    \caption{\textbf{Sparse scaling for few-shot inference}. \textbf{Left:} \cite{du2021glam} measures the few-shot inference performance on TriviaQA, demonstrating consistent gains of sparse MoE models over dense models up to 137B parameters. Each label, such as 8B/64E, says how many parameters per input are used (8B) and how many experts (64E). \textbf{Right:} BigBench \citep{srivastava2022beyond} studied the few-shot scaling properties on a larger set of 161 contributed JSON tasks to confirm improvements of sparse expert models over their FLOP-matched dense counterparts.}
    \label{fig:few_shot_scaling}
\end{figure}

\cite{du2021glam} presented the scaling of sparse GLaM models ranging from 1B-64B FLOPs using 64 experts per sparse layer. 
GLaM achieved state-of-the-art results, outperforming the 175B parameter GPT-3 \citep{brown2020language} model in zero and one-shot performance, while using 49\% fewer FLOPs per token at inference and 65\% lower power (left plot in Figure \ref{fig:few_shot_scaling}).
In another example of sparse models performing well on few-shot inference, the BIG-Bench \citep{srivastava2022beyond} collaboration measured a 2$\times$ improvement of sparse over dense models on the 161 contributed JSON tasks (right plot in Figure \ref{fig:few_shot_scaling}).

\begin{figure}[ht!]
    \centering
    \includegraphics[width=0.6\textwidth]{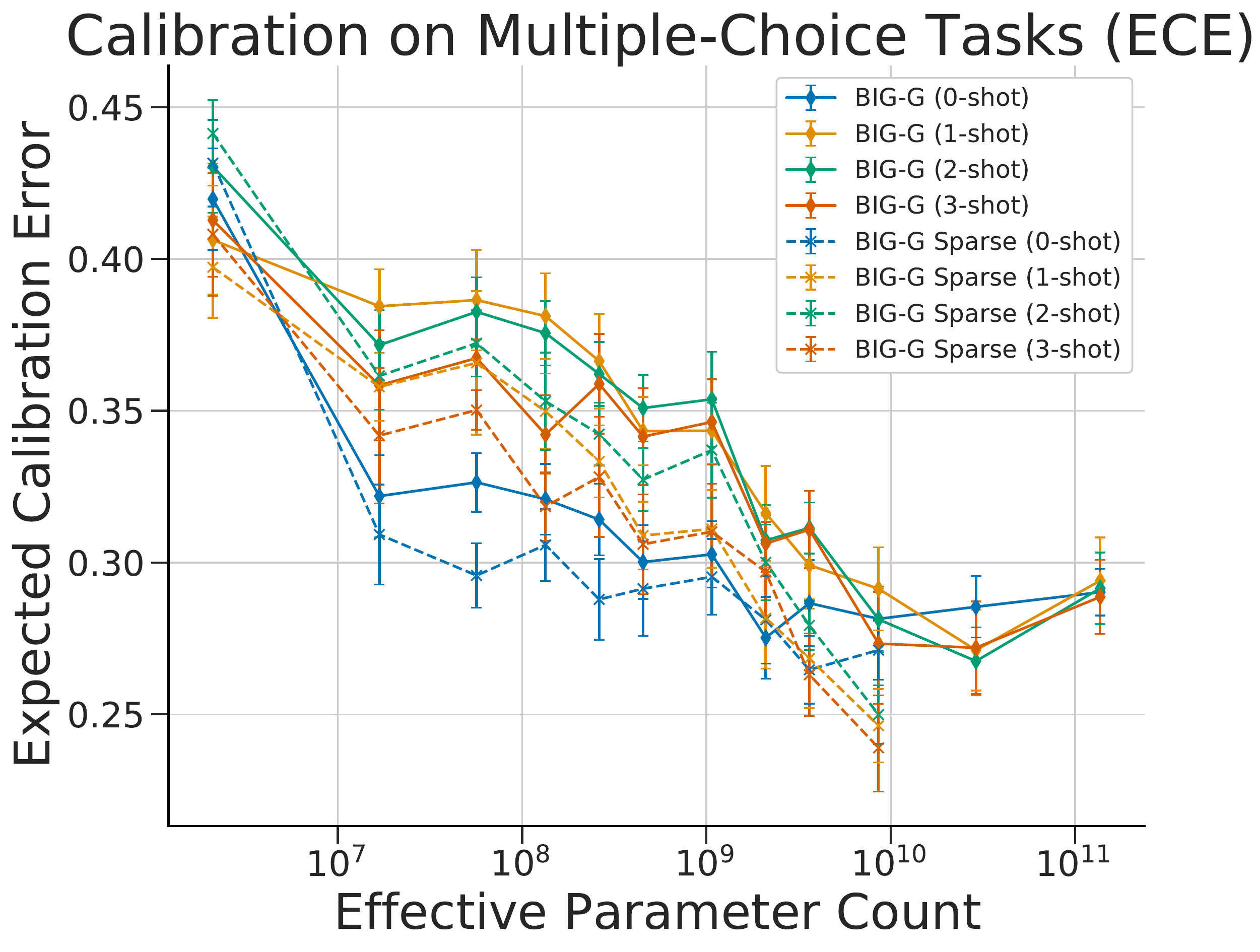}
    \caption{\textbf{Sparse model calibration}. The Expected Calibration Error improves with scale for both dense and spare models. However, sparse models exhibit significantly better calibration and roughly match the calibration of a 10$\times$ larger dense model. Figure is reproduced from \cite{srivastava2022beyond}.}
    \label{fig:calibration}
\end{figure}

Finally, \cite{srivastava2022beyond} studied the \emph{calibration} of sparse models on the multiple choice BIG-Bench tasks.
Calibration measures the degree to which the probability of a prediction matches the probability of being correct.
This work measured calibration by the Expected Calibration Error \citep{naeini2015obtaining} which is the absolute deviation between the predicted probability and average accuracy, after binning examples by their predicted probability.
While the calibration improves for both larger dense and sparse models (Figure \ref{fig:calibration}), the sparse models were found to match the calibration of a dense model using 10$\times$ more FLOPs.

\subsection{Scaling the Number, Size and Frequency of Expert Layers}

Several important hyperparameters, beyond those in a dense Transformer, govern the scale of sparse expert models including, 1) the expert count, 2) the size of each expert, and 3) the frequency of expert layers.
The decisions can have significant implications to the upstream and downstream scaling.

Many earlier works scaled to thousands of relatively small experts per layer, which has produced excellent pre-training and translation quality \citep{shazeer2017outrageously,lepikhin2020gshard,fedus2021switch}.
However, the quality of sparse models is disproportionately reduced under domain shift \citep{artetxe2021efficient} or when fine-tuning on different task distributions \citep{fedus2021switch}.
The state-of-the-art sparse models for few-shot inference (GLaM \citep{du2021glam}) and for fine-tuning (ST-MoE \citep{zoph2022designing}) use only up to 64 larger experts -- a better balance of computation and parameters.
As a result of the increased expert dimensions, these models require specific system-level sharding strategies over accelerators to run efficiently \citep{du2021glam,rajbhandari2022deepspeed}.

Next, we recap the current conventions around the frequency of expert layers. 
Usually, sparse models are constructed by beginning with a dense model and either inserting or substituting sparse expert layers at either a fixed interval or heuristically. As an example, \cite{rajbhandari2022deepspeed} put more sparse layers near the final layers of the network.
In the Transformer, the most common approach is to replace every other Feed-Forward Network (FFN) layer \citep{lepikhin2020gshard,du2021glam,artetxe2021efficient,rajbhandari2022deepspeed}, that is, substitute with a frequency of 0.5.
However, other frequencies have been used, including every-fourth-layer (0.25) in \cite{zoph2022designing} and every-layer (e.g. 1.0) in \cite{fedus2021switch}.
Finally, a frequency of 0.5-1.0 is recommended by \cite{clark2022unified}.

Ultimately, the answer to the question of optimal hyperparameters depends on the application and the hardware system specifications.
Prior work demonstrated strong pre-training and translation results with a high number of experts \citep{shazeer2017outrageously,lepikhin2020gshard}, whereas, the best performing models under transfer have used fewer, larger experts \citep{du2021glam,zoph2022designing,mustafa2022multimodal}.
Further, these decisions are highly hardware-dependent.
Due to the added \texttt{all2all} communication costs to implement routing, networks with slower interconnect speeds may find fewer expert layers is optimal on a time-basis to a certain quality.
A simulation of the compute, memory, and communication properties of a distributed system would significantly aid practitioners to more quickly determine optimal settings, without costly trail-and-error launches.

We note that this analysis and trade-offs are for experts-as-a-layer approach \citep{eigen2013learning}.
In contrast, the Branch-Train-Merge (BTM) approach to experts \citep{li2022btm} is ``embarrassingly parallel`` in that each expert is a fully formed language model, trainable independently and asynchronously, without the expensive communication costs.
Therefore, this approach and others following-suit have completely different scaling characteristics with the number of experts.

\section{Routing Algorithms} \label{sec: routing}
The \emph{routing algorithm}, a key feature to all sparse expert architectures, determines where to send examples.
This area has been studied extensively, including counter-intuitive methods that use fixed, non-learned routing patterns \citep{roller2021hash}.
Typically the naive routing decision is non-differentiable because it makes a discrete decision of which experts to select.
The problem of expert selection can be recast as a Bandit problem and several works have used reinforcement learning to learn the selection \citep{bengio2016conditional,rosenbaum2017routing,rosenbaum2019routing,clark2022unified}.
\cite{shazeer2017outrageously} proposed a differentiable heuristic that side-stepped reinforcement learning challenges.
Rather than routing the example to the chosen expert and proceeding, the output of the expert computation is weighted by the \emph{probability} of choosing it (Equation \ref{eqn: moe_layer}).
This produces a gradient to the router since the probability of choosing the expert is differentiable.
In contrast to a Bandit approach, where only one expert might be chosen, \cite{shazeer2017outrageously} conjectured that it was necessary to route to the top-$k$ experts with $k > 1$.
The intuition was two or more experts on the same example allowed the network to compare and to optimize the relative performance.
\cite{lepikhin2020gshard} later adapted the same routing algorithm to the Transformer architecture, yielding state-of-the-art machine translation results.
However, \cite{fedus2021switch} demonstrated that top-$1$ routing can achieve competitive results, corroborated by later work \citep{clark2022unified}.

\subsection{Routing Taxonomy}
\begin{figure}[ht!]
    \centering
    \includegraphics[width=0.65\textwidth]{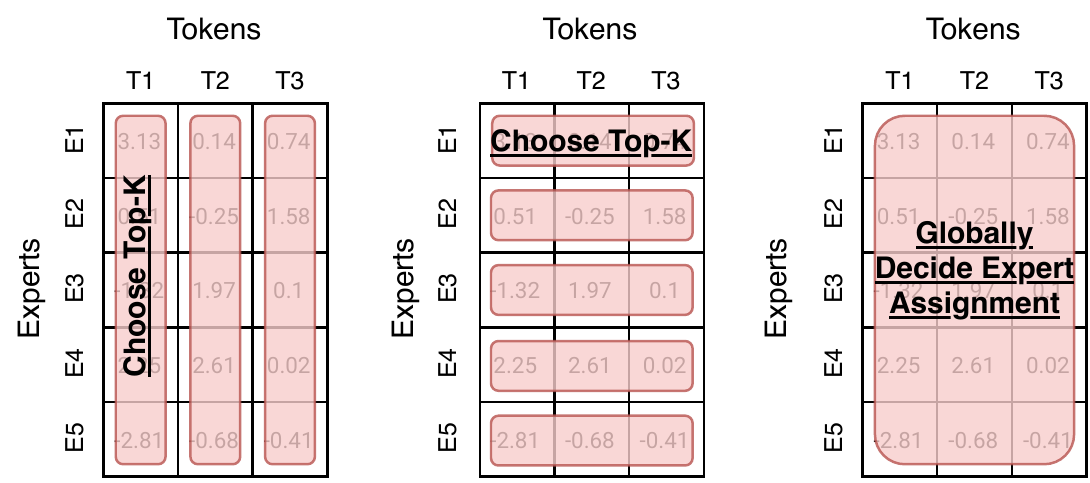}
    \caption{\textbf{Three common classes of routing algorithms}. We illustrate three methods with an \texttt{Experts} $\times$ \texttt{Tokens} activation matrix obtained through the process explained in Figure \ref{fig: router_math}. \textbf{Left:} ``Choose Top-$k$'' along the \texttt{Experts} axis includes the standard top-$k$ routing algorithm \citep{shazeer2017outrageously,lepikhin2020gshard}. \textbf{Center:} ``Choose Top-$k$'' along the \texttt{Tokens} axis are routing algorithms such as \cite{zhou2022mixture}. \textbf{Right:} ``Globally Decide Expert Assignment'' routing algorithms such as BASE layer \citep{lewis2021base,clark2022unified}.}
    \label{fig: router_scores}
\end{figure}

One way to understand many routing algorithms is to analyze the matrix of routing scores (i.e. the router scores from Figure \ref{fig: router_math}).
As a demonstrative example, we use a natural language sparse expert model.
Figure \ref{fig: router_scores} shows the un-normalized router scores computed for three tokens (columns) routed across five experts (rows).
Each value is produced by the dot product of the token embedding and the expert embedding (from the router weights). 
Once the scores are computed, there are a variety of ways to determine which experts should get which tokens.
We highlight three common categories: 1) each token chooses the top-$k$ experts, 2) each expert chooses the top-$k$ tokens, and 3) globally determine what tokens should go to each expert (and not use a greedy approach).
This taxonomy also further suggests yet to be explored routing algorithms.
One example is an algorithm that benefits by looking both horizontally and vertically at the router scores, but without incurring the cost of looking globally.
A token could first choose what experts it wants to go to, then based on this information, each expert could choose what tokens it wanted.

\begin{figure}[ht!]
    \centering
    \includegraphics[width=\textwidth]{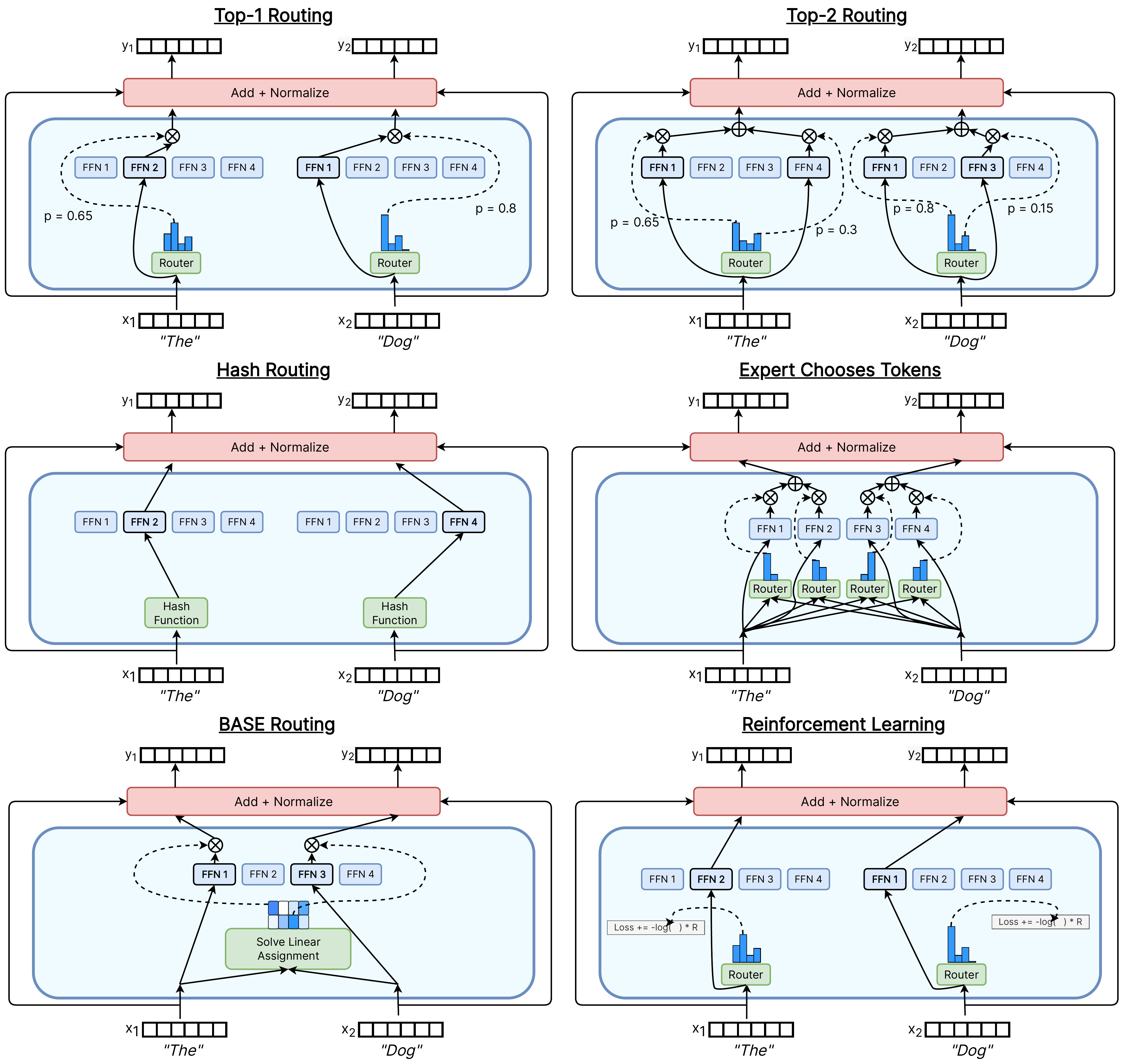}
    \caption{\textbf{Visualization of six different routing algorithms.} Each diagram is of a Transformer sparse expert model with four experts (feed-forward networks) routing two tokens: ``The'' and ``Dog''.}
    \label{fig: routing_algorithms}
\end{figure}

\paragraph{Each token chooses the top-$k$ experts.}This class of routing algorithms have each token choose the top-$k$ experts to be sent to. This is the original routing top-$2$ formulation proposed in \cite{shazeer2017outrageously} and used in \cite{lepikhin2020gshard} that achieves state-of-the-art machine translation results.
\cite{fedus2021switch,rajbhandari2022deepspeed} used top-1 routing with success.
\cite{clark2022unified} proposed a reinforcement learning routing algorithm that used top-1 routing. 
However, instead of scaling the output of the expert computation by the router probability, they use REINFORCE \citep{williams1992simple} with the reward being the negative cross entropy of the predicted token. Figure \ref{fig: routing_algorithms} depicts the top-1, top-2 and reinforcement learning routing algorithms.

\cite{yang2021m6t} introduced an extension of top-1 routing by using expert prototyping to split experts into different groups and then applied $k$ top-1 routing procedures. \cite{nie2021dense} begins routing as a soft gating function where all experts are trained (e.g.\@ a dense model) and anneals down to the standard top-1 routing algorithm. This approach (DTS-Gate) improves over Switch Transformer \citep{fedus2021switch} on OpenWebText pre-training.  \cite{dua2021tricks} proposes a similar approach of first training a dense model where each inputs goes to every expert and then adapting it to be sparse. \cite{hazimeh2021dselectk} proposes DSelect-k, which is a smooth version of the top-$k$ routing algorithm that improves over standard top-$k$ routing. \cite{rajbhandari2022deepspeed} designs PR-MoE which uses top-2 routing, but each token is sent to a shared dense layer and a single expert of its choosing (instead of two experts).

\cite{riquelme2021scaling} introduces an improvement for top-$k$ routing named Batch Prioritized Routing (BPR) for ViT image classification models. MoE models use fixed batch sizes per expert, which can cause tokens to overflow if there is not enough capacity. If a token overflows then no computation will be applied to that token at that given expert. In top-$k$ routing, priority of which tokens to not drop at an overflowed expert is given to the tokens sent earlier in a sentence/batch. BPR instead prioritizes inputs that have higher routing scores. This is relevant for ViT models as there is no autoregressive nature to the inputs, so all inputs can see each other. In language there is typically a left-to-right ordering of the inputs, which could in theory allow the model to cheat during training. \cite{zoph2022designing} found BPR routing to be helpful for MoE language models. \cite{kim2021scalable} proposes randomizing the prioritization of the tokens in the sequences to make sure the routing is not biased towards early tokens in the sentence.

\paragraph{Static routing.}Most routing algorithms dynamically learn the routing decisions while training, but this can be statically determined before training begins. 
Dynamic routing algorithms typically operate on the internal input representations within the network, so the routing decisions take into account the current token and previous inputs to the model (usually through the self-attention layer). 
Most routing algorithms are dynamic, but a notable example of a static routing algorithm is is Hash Layers from \cite{roller2021hash}.
This work shows \emph{random} fixed routing by hashing the input token led to competitive performance with learned routing.
Load balancing is achieved by choosing hash functions before training that balances batches of tokens.
A depiction of Hash Layers can be found in Figure \ref{fig: routing_algorithms}. 

\paragraph{Each expert chooses the top-$k$ tokens.}
Instead of each token choosing what experts to be sent to, \cite{zhou2022mixture} flips this and has each expert choose what tokens it wants routed to it. This alleviates the need for auxiliary load balancing losses to be added during training or for linear assignment algorithms. Now each expert will always have the same amount of tokens, although some tokens might not get sent to any expert or some tokens might get sent to all of them. Empirically this algorithm performs well and has an adaptive computation interpretation where the model can implicitly apply more computation to certain tokens.

\paragraph{Globally determine what tokens should go to each expert.}BASE layers \citep{lewis2021base} treats token routing as a linear assignment problem. It aims to route a fixed number of tokens to each expert and maximize the scores from the routing matrix. Since the tokens per processor are highly correlated as they come from the same sentences, tokens are randomly shuffled around before locally solving the linear assignment problem on each device. This shuffling introduces two additional communication primitives (\texttt{all2all}) in both the forward and backward pass. \cite{clark2022unified} proposes their own variant of BASE layer (S-BASE) that uses an optimal transport formulation.

\paragraph{Other routing algorithms.} Some routing algorithms do not neatly fall into the above three categories. \cite{zuo2021taming} introduced THOR, an algorithm which \emph{randomly} selects two experts for each input during training and inference and found improvements of 2 BLEU points over standard MoE models.
\cite{gururangan2021demix} proposes DEMix, which explicitly has different experts for different pre-training domains (e.g. law, medical, etc.). 
Experts can then be selected by doing domain matching on the inputs.
\cite{fan2021beyond} uses explicit language-specific sublayers where input tokens can be deterministically routed based on their language.
This avoids needing dynamic routing algorithms. \cite{ma2018modeling} introduces a multi-gate routing algorithm where each task gets it own unique gating function.

\subsection{Load Balancing}
Most routing algorithms handle load balancing by adding an auxiliary loss during training to encourage equal amounts of tokens getting sent to the different experts \citep{shazeer2017outrageously}. Some routing algorithms handle load balancing through their design: BASE Layers \citep{lewis2021base} solves a linear assignment problem that enforces an equal number of tokens going to each expert as part of the problem statement. S-BASE from \cite{clark2022unified} follows a similar protocol, but solves the assignment problem using optimal transport. \cite{nie2021dense} starts by training a Mixture-of-Expert model where all tokens get sent to each expert, but over time adapts the network to do top-1 routing. This algorithm doesn't need load balancing as the network naturally learns to the specialize the expert representations over training.

\section{Sparse Expert Models Across Domains}

Sparse expert and MoE models were introduced and popularized in natural language processing (NLP). 
This domain was a natural fit for large models which benefit from the easy availability of trillions of tokens and the strong self-supervised algorithms of next word prediction and masked language modeling. 
However, the impact of these models is quickly spreading to other domains including computer vision, speech recognition and multi-modal applications.
The spread of techniques from NLP to other domains has been accelerated because the Transformer has been rapidly adopted in other domains and modalities. 
Some examples are image classification \citep{dosovitskiy2020image}, object detection \citep{carion2020end}, recommendation systems \citep{chen2019behavior}, speech recognition \citep{dong2018speech,nakatani2019improving,gulati2020conformer}.

Across various domains, the sparse architectures and algorithms stay roughly the same, but what is routed to the experts is different. 
Table \ref{tab: moe_inputs} shows the different sparse layer inputs for a variety of different domains.
\begin{table}[ht!]
    \small
    \centering
    \begin{tabular}{c|ccc|p{6cm}}
    \toprule
        \textbf{Domain} & \textbf{Input Representation} \\
        \midrule
        NLP & Word, subword, or sentence \\
        Vision & Image patch \\
        Speech & Spectrogram \\
        Multimodal & Word or image patch \\
    \bottomrule
    \end{tabular}
    \caption{\textbf{Inputs to sparse models in different domains.} The input is used to determine what expert to route to and is what the MoE layer will apply compute to.}
    \label{tab: moe_inputs}
\end{table}

\subsection{Natural Language Processing}

Initially \cite{shazeer2017outrageously} introduced the Mixture-of-Expert layer for LSTM language modeling and machine translation. The layers were inserted between the standard layers in the LSTM model. Follow up works are now based around Transformers, and the expert layers typically replace the dense layers. 

\cite{lepikhin2020gshard} first introduced the MoE layer into Transformers and studied it in the context of machine translation. They achieved state-of-the-art translation results across 100 different languages when scaling up to 2048 experts per expert layer. \cite{fedus2021switch} later created a sparse 1.6T parameter language model that achieved state-of-the-art pre-training quality.
They also studied using sparse layers to produce the $q$/$k$/$v$ activations in the Self-Attention layers, but found this technique to be more unstable. \cite{lee2022sparse} introduces the Fast Sparse Mixer, which is an encoder only model that achieves 89\% training and 98\% inference speedups over BERT \citep{devlin2018bert}.

Recently, there has been a flurry of MoE research on a variety of different topics in the NLP domain.
As an example, prior MoE architectures in the NLP domain acted on the word or byte-pair level. \cite{kudugunta2021beyond} instead had an MoE architecture route at the task or sentence level, which allows for more efficient inference and serving. 
This was studied in the context of machine translation where sentences would be routed based on what language they were translating into.

New results have been able to push state-of-the-art on few-shot inference and fine-tuning benchmarks. \cite{du2021glam} trained a MoE decoder-only language model and achieved state-of-the-art few-shot results, while requiring only 1/3 the compute needed to train GPT-3. \cite{zoph2022designing} introduced ST-MoE, a sparse encoder-decoder model that achieves state-of-the-art on a large set of reasoning and generation tasks including SuperGLUE, ARC Easy/Challenge, XSum, CNN-GM, Web-QA, ANLI, and Winogrande. 
ST-MoE outperforms PaLM-540B \citep{chowdhery2022palm} when fine-tuning on SuperGLUE, while using roughly 20$\times$ less pre-training FLOPs and 40$\times$ less inference FLOPs.

\subsection{Computer Vision}

Due to the universality of Transformers (e.g. ViT \citep{dosovitskiy2020image}), applying improvements to the MoE architecture across domains has been fruitful. \cite{riquelme2021scaling} created a vision MoE model by adding MoE layers into the ViT architecture. Their model, V-MoE, was applied to image classification and was able to use just half the amount of inference compute while matching the performance of prior state-of-the-art architectures. \cite{lou2021sparse} introduces a sparse MoE MLP model for image classification based on the MLP-Mixer architecture \citep{tolstikhin2021mlp}. Their MoE variant achieved better image classification performance on ImageNet and CIFAR compared to its dense  counterpart.

\cite{wu2022residual} improved the efficiency of training MoE models through their proposed Residual Mixture-of-Expert layer. This architecture achieves 30\% reduced training cost and comparable quality to standard MoE models on both segmentation and object detection. \cite{hwang2022tutel} implements an efficient framework and adaptive parallelism strategy for MoE layers. To benchmark their system, they add MoE layers to the Swin Tranformer V2 architecture \citep{liu2022swin} for image classification and object detection. Their MoE variant achieves 1.5x-2x speedups in both training and inference over the previous MoE implementation. \cite{aljundi2017expert} uses expert models in a continual learning setup where they add new experts over time and demonstrate improvements on image classification and video prediction.
\cite{caccia2021anytime} dynamically increases the expert count over the course of training for image classification models on CIFAR and MNIST. \cite{ramachandran2018diversity} studies how depth and architectural diversity impacts sparse expert model performance and achieves gains on image recognition. \cite{kirsch2018modular} develops an end-to-end algorithm for doing conditional computation based on the input data that achieves strong image classification performance.

\subsection{Speech Recognition}

SpeechMoE \citep{you2021speechmoe} uses MoE Transformer models for speech recognition and achieves strong character error rate improvements across four datasets. They use novel auxiliary losses to promote sparsity and introduce a new routing architecture. SpeechMoE2 \citep{you2022speechmoe2} further improves on the SpeechMoE's results by making a new routing algorithm that adds in new auxiliary information for making routing decisions. \cite{kumatani2021building} yields improvements for multi-lingual speech recognition by using MoE layers in two different types of Transformer speech architectures: seqeuence-to-sequence and transducers. 

\subsection{Multimodal and Multi-Task}
\cite{mustafa2022multimodal} does multimodal learning by training a MoE model (LIMoE) that takes as input both images and text and learns using a contrastive loss similar to CLIP \citep{radford2021learning}. The MoE layer can route both the image patches and the word tokens to the available experts. The model outperforms CLIP when using a comparable training strategy, and when scaled further matches state-of-the-art methods.

\section{When to Use a Sparse Versus Dense Model}
A common question is if you are given a fixed compute or FLOP budget (e.g. 100 GPUs for 20 hours), what type of model should you train to achieve the best performance?
Many prior works show that sparsity is better than a dense model for this type of setup \citep{shazeer2017outrageously,lepikhin2020gshard,fedus2021switch,du2021glam,artetxe2021efficient,lewis2021base}.
Given all the strong state-of-the-art results using sparse models, why should you ever not use a sparse model over a dense model?

Sparse models are a generalization of a dense model; a sparse model with a single expert is \emph{roughly} a dense model.
Fundamentally, sparse models allow to vastly increase the number of parameters in a model by increasing the number of experts, while keeping the FLOPs per example approximately constant.
This can be good or bad depending on the setup and how the model is going to be used later.

At a high level, sparsity is good when you have many accelerators (e.g. GPU/TPU) to host all the additional parameters that comes when using sparsity.
Typically models are trained using data-parallelism where different machines will get different slices of the training/inference data. 
The machines used for operating on the different slices of data can now be used to host many more model parameters.
Therefore, sparse models are good when training with data parallelism and/or have high throughput while serving: training/serving on many machines which can host all of the parameters. 

Using sparsity requires careful consideration for how the model will be used in downstream usage too.
If there are lots of machines to pre-train a model, but a lot less for fine-tuning or serving, then the amount of sparsity (e.g. the number of experts) should be tailored to fit the amount of memory available in the downstream use cases.
This is often a practical design consideration used in the literature.

On a per \emph{parameter} basis, sparse models will always look comparatively worse to dense models. 
Assuming that all parameters are kept in the accelerators memory, this is a similar requirement to seeking the best model that can fit onto a certain hardware size (e.g. 4 GPUs), where again a sparse model will be a worse option than a dense one.
As mentioned above, sparse models are a great fit when you have the ability to either be training or serving on many machines in parallel in order to host the additional model parameters from the experts. 

All hope is not lost for sparse models in memory restrictive settings though. 
\cite{fedus2021switch} shows that sparse models work well with as few as two experts, which requires limited additional memory.
New research also allows for overcoming GPU/TPU memory shortages by dynamically swapping model memory between the CPU and GPU (see Section \ref{sec: hardware} for more details). 
Other approaches for reducing the memory footprint of sparse models are discussed in Section \ref{sec: inference}.

\section{Sparse Model Training Improvements}

Sparse models often have different dynamics than dense models and benefit from different training and fine-tuning methodologies. 

\subsection{Instability}
Sparse models have frequently been reported to be more unstable, meaning the loss diverges and increases \citep{lepikhin2020gshard,fedus2021switch,zoph2022designing,mustafa2022multimodal}. 
Instabilities also appear more often at larger model scales. 
\cite{lepikhin2020gshard} encountered training instability using \texttt{bfloat16} activations with a 1 trillion parameter model. 
\cite{fedus2021switch} encountered instabilities in their highest-compute Switch-XXL model. 
\cite{zoph2022designing} encountered instabilities in their largest models, especially in the multi-lingual setting. 
\cite{mustafa2022multimodal} observed increased instability when doing multi-modal training on both images and text.

Much research has been done to improve the training dynamics of sparse models. \cite{lepikhin2020gshard} noted that the largest model instabilities can be fixed by training the models using higher precision (\texttt{float32}), but comes at the cost of more memory usage and slower training. \cite{fedus2021switch} recommended using a lower weight initialization scale and casting only a specific subset of the routing network to higher precision for better model stability/training. \cite{du2021glam} skips batches of data that have any NaNs or Infs in the gradients and also restarts the model from an earlier checkpoint when any training divergences occur. \cite{artetxe2021efficient} propose a smarter initialization of the expert layer to account for the reduced batch size of the expert weights. Since each expert will have a batch size of $\frac{B}{E}$\footnote{$B$ is the tokens in the batch and $E$ is the number of experts. This assumes top-1 routing, but similar analysis holds for the other routing variants.} the authors propose scaling the gradients of the expert layer by $\frac{1}{\sqrt{E}}$. \cite{zoph2022designing} introduced the router z-loss to improve both the model instability and also quality. This auxiliary loss aims to reduce floating point roundoff errors by encouraging the logits going into the router function to remain small over the course of training. \cite{mustafa2022multimodal} extensively studied many techniques to fix model instability, and used a combination of different approaches including the router z-loss and two novel entropy losses.

\subsection{Transfer to New Distributions}
Several research papers, especially at larger scales, note that MoE models transferred to new domains (as in fine-tuning) lags their dense counterparts. 
\cite{fedus2021switch,narang2021transformer} compared the pre-training perplexity versus fine-tuning performance for dense and sparse models. They noticed for a given pre-training perplexity, sparse models were fine-tuning worse on reasoning tasks, but better on knowledge heavy tasks. 
In addition to worse out-of-domain language modeling performance, \cite{artetxe2021efficient} observed worse fine-tuning compared to dense models on multiple tasks including HellaSwag, PIQA and Winogrande.

Several different ways have been proposed to help address the fine-tuning issues. One is to scale models by having more FLOPs compared to more sparsity (e.g. fewer experts, but make them larger). 
\cite{fedus2021switch} trained a 1.6T parameter model with 2048 experts, but it only had as many FLOPs as a 2B dense model.
Conversely, the model with the best fine-tuning performance had only 128 experts, but the amount of FLOPs as a 11B dense model. 
Trading off less sparsity for more FLOPs when scaling a model is a simple way to ensure better fine-tuning performance. \cite{zoph2022designing} noticed that the optimal fine-tuning hyperparameters (e.g. learning rate and batch size) can be dramatically different for dense and sparse models.
Using the best hyperparameters for dense models on sparse models can mask any of the sparsity pre-training improvements -- therefore, an independent hyperparameter study is beneficial.

\subsection{Inference} \label{sec: inference}
By design, sparse expert models have many more parameters than their dense counterparts.
While the computation done is still relatively low, the memory footprint can be a burden.
Therefore, some research has focused on reducing the number of parameters needed at inference-time to ease serving requirements. 
\cite{kudugunta2021beyond} routes at the \emph{task} level instead of the word or token level for machine translation. 
This allows for more efficient inference because the subset of weights only for the needed tasks are required. 
\cite{kim2021scalable} prunes away experts at inference to reduce the memory footprint of the model.
Two different methods are used for pruning: randomly selecting a subset of the experts and choosing the experts with the highest utilization at inference time. \cite{fedus2021switch} distill large sparse models into smaller dense models for language modeling and fine-tuning. \cite{rajbhandari2022deepspeed} studies distillation of sparse models for language modeling by reducing the depth in the expert layers of the network. \cite{rajbhandari2022deepspeed} implements an optimized version of MoE into the DeepSpeed framework that results in 7.3$\times$ faster inference latency than existing frameworks.

\section{Interpretability}
Sparse expert models more naturally lend themselves to interpretability studies because each input is processed by an identifiable, discrete subset of the model weights (i.e. the chosen experts).
Therefore, instead of the daunting task of interpreting possibly trillions of floating point numbers, one can instead read off a small discrete set of integers corresponding to which expert the input was sent.
\cite{shazeer2017outrageously} conducted preliminary studies into expert specialization for the encoder of their 2048 MoE layer on the WMT '14 EnFr machine translation task. They identified three experts, one with a specialization of words around innovation, the second which processed the article ``a'', and a third which was routed synonyms of speed.

Later, more extensive analyses were conducted by \cite{lewis2021base} on Transformer-based architectures.
\cite{lewis2021base} conducted a study where they tracked the most frequent prior input token when the expert was selected.
This revealed specialization in quantities, numbers, possessives, subword fragments, and clusters of related verbs, nouns and adjectives, with selected results presented in Table \ref{tab:base_expert_specialization}.

\begin{table}[h!]
\small
\begin{center}
\renewcommand{\arraystretch}{1.2}
\begin{tabular}{l|l}
    \toprule
    \textbf{Expert} & \textbf{Top-5 preceding tokens} \\
    \midrule
    5 & year, years, billion, millions, tonnes \\
    9 & electronic, local, public, national, outdoor \\
    34 & to, will, should it, may \\
    42 & two, 50, 1, 80, 000 \\
    62 & work, started, involved, working, launched \\
    72 & is, was, be, been, were \\
    74 & going, go, come, back, return \\
    101 & B, T, W, H, k \\
    \bottomrule
\end{tabular}
\vspace{-0.2cm}
\caption{
\textbf{Expert specialization based on preceding context in BASE Layers}.
We reproduce a portion of table of \cite{lewis2021base}, presenting the most frequent preceding top-five tokens for the selected experts. This example shows experts specializing in punctuation, conjunctions \& articles, verbs, visual descriptions, proper names, counting \& numbers.}
\label{tab:base_expert_specialization}
\end{center}
\vspace{-0.15cm}
\end{table}

\cite{zoph2022designing} trained an encoder-decoder Transformer and finds similar patterns in the encoder, including experts that specialize in a shallow way, such as over articles (e.g. ``a'', ``the'').
Table \ref{tab:encoder_expert_specialization} reproduces a portion of the observed specializations of \cite{zoph2022designing}.
Those studies further found expert specialization in punctuation, numbers, proper names, verbs, colors and special mask tokens used for the pre-training objective.

\begin{table}[h!]
\small
\begin{center}
\renewcommand{\arraystretch}{1.2}
\begin{tabular}{l|l|l}
    \toprule
    \textbf{Expert specialization} & \textbf{Expert position} & \textbf{Routed tokens} \\
    \midrule
    \textbf{Sentinel tokens} & Layer 1 & been \textless extra\_id\_4\textgreater \textless extra\_id\_7\textgreater floral to \\
    & & \textless extra\_id\_10\textgreater \textless extra\_id\_12\textgreater \textless extra\_id\_15\textgreater \\
    & & \textless extra\_id\_17\textgreater \textless extra\_id\_18\textgreater \textless extra\_id\_19\textgreater ... \\
    \midrule
    \textbf{Punctuation} & Layer 2 & , , , , , , , , , - , , , , , ). )  \\
    & Layer 6 &  , , , , , : . : , \& , \& \& ? \& - , , ? , , , . \textless extra\_id\_27\textgreater \\
    \midrule
    \textbf{Conjunctions and articles} & Layer 3 & The the the the the the the the the The the the \\
    & Layer 6 & a and and and and and and and or and a and . \\
    \midrule
    \textbf{Verbs} & Layer 1 & died falling identified fell closed left posted lost felt \\
    & & left said read miss place struggling falling signed died \\
    \midrule
    \textbf{Visual descriptions} & Layer 0 & her over her know dark upper dark outer \\
    \textit{color, spatial position} & & center upper blue inner yellow raw mama \\
    & & bright bright over open your dark blue \\
    \midrule
    \textbf{Proper names} & Layer 1 & A Mart Gr Mart Kent Med Cor Tri Ca Mart \\
    & & R Mart Lorraine Colin Ken Sam Ken Gr Angel A \\
    \midrule
    \textbf{Counting and numbers} & Layer 1 & after 37 19. 6. 27 I I Seven 25 4, 54 I two dead we \\
    \textit{written and numerical forms} & & Some 2012 who we few lower each \\
    \bottomrule
\end{tabular}
\vspace{-0.2cm}
\caption{
\textbf{Encoder expert specialization in ST-MoE}.
We reproduce a table of \cite{zoph2022designing} demonstrating expert specialization in punctuation, conjunctions \& articles, verbs, visual descriptions, proper names, counting \& numbers.}
\label{tab:encoder_expert_specialization}
\end{center}
\vspace{-0.15cm}
\end{table}

But a deeper analysis of the full encoder-decoder ST-MoE architecture found clearer evidence of specialization in the encoder, rather than the decoder. 
This warrants further study into the value and positioning of expert layers.
Lack of evident specialization may either signal a difficult to discern patterns or no useful patterns.

\begin{figure}
    \centering
    \includegraphics[width=0.95\textwidth]{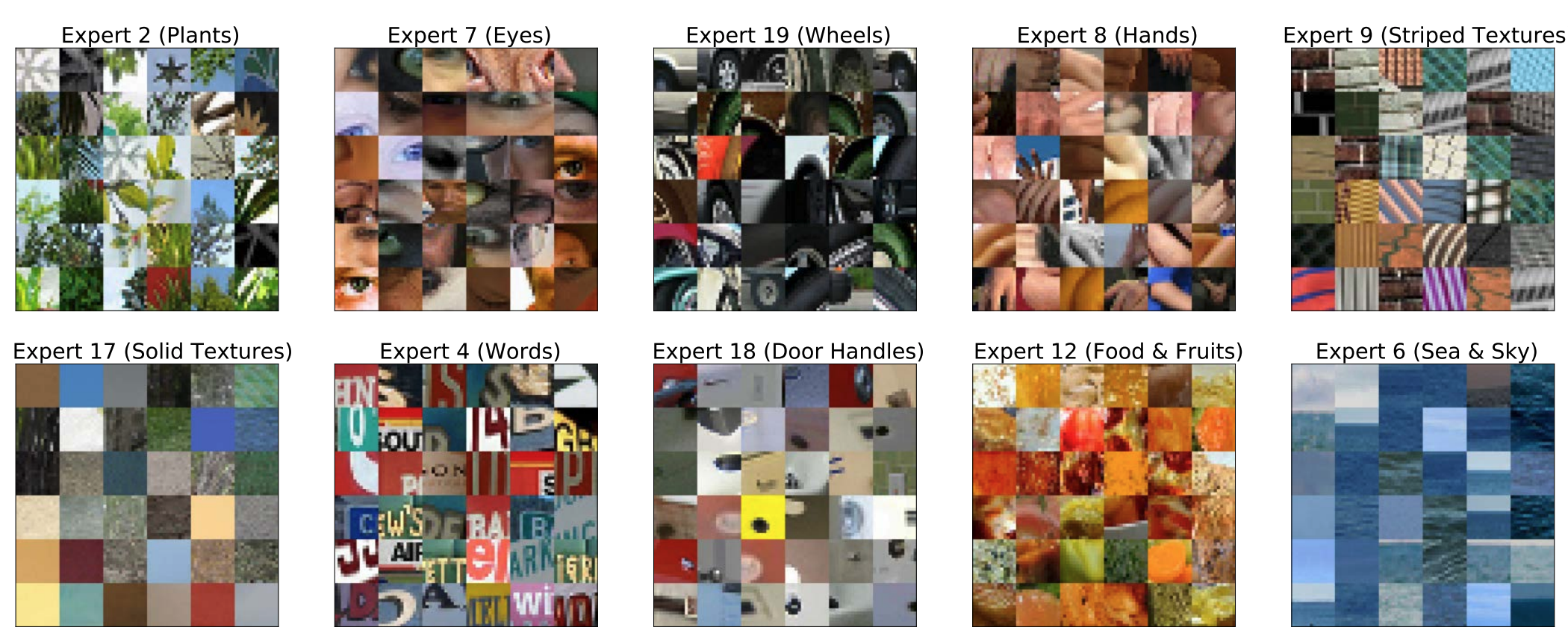}
    \caption{\textbf{Visual 
    expert specialization in LIMoE}. We reproduce a figure from \cite{mustafa2022multimodal} which finds expert specialize in patches of textures (solid and striped), natural objects (plants, hands, eyes), and man-made objects (wheels, door handles, words).}
    \label{fig:image_experts}
\end{figure}

Interpretability of sparse expert models has not only been limited to text.
One example is LIMoE \citep{mustafa2022multimodal}, a multi-modal model that was observed to learn experts that specialize in textual and visual data, including patches of textures, plants, eyes, and words (Figure \ref{fig:image_experts}).
As in the text based models, the complexity of the specialization varies significantly.
For instance, text-based experts were found to span simple objectives like processing the article ``a'' up to more complicated concepts like past-tense verbs. 
Similarly, in multi-modal models, the sophistication of expert specialization varies to concepts as simple as a basic textures up to high-level objects such as wheels or door handles.

Finally, we highlight one significant limitation of these interpretability approaches. 
These consider the tokens or patches arriving to each expert in a narrow way.
Specifically, the initial embedding of a word or of a patch incorporates contextual information from surrounding data (Transformers do this through self-attention or encoder-decoder attention).
Therefore, more nuanced specialization may be missed by these heuristic techniques. 
More careful and thorough interpretabilty work will be needed to better understand sparse expert models.
The release of spare expert model checkpoints by \cite{artetxe2021efficient} and \cite{fedus2021switch} allows broader groups to analyze and explain these dynamics.

\section{Future Directions and Conclusions}
Even though sparse expert models and Mixture-of-Expertss date back to at least the early nineties -- many questions remain.
We conclude our review with a conjecture of promising areas of future work, specifically highlighting the intersection with two recent developments (adaptive computation and retrieval methods) and our parting thoughts.

\paragraph{Adaptive Computation. }
Adaptive computation is the idea that different inputs to a machine learning system may use differing amounts of computation (i.e. the amount or the type of compute is adapted on-the-fly). 
Sparse models build on the mirrored idea: each input uses the same amount of computation, but potentially with different parameters. 
However, these techniques are not mutually exclusive; some routing algorithms (Section \ref{sec: routing}) allow for adaptive computation by sending a token to a variable number of experts \citep{riquelme2021scaling,zhou2022mixture}. 
Still future models may benefit by combining other adaptive computation techniques -- as an example, in addition to choosing which expert, a network might choose the number of layers to use, as well \citep{schuster2022confident}.
\emph{Heterogeneous} expert layers also are a natural fit for adaptive computation.
Most sparse models uses experts of the same type and size for simplicity and efficiency on modern hardware.
But by allowing experts to differ in size (e.g. in depth or width), the routing decision will then result in differing amounts of computation.
New software systems, such as Pathways \citep{dean2021pathways}, will help facilitate efficient implementations of these heterogeneous architectures and algorithms on modern hardware.

\paragraph{Retrieval Methods. }
Retrieval mechanisms effectively expand the capacity of models by allowing them to dynamically access information beyond the current context or what is stored in the parameters \citep{khandelwal2019generalization,guu2020retrieval,borgeaud2022improving}.
Sparse expert models and retrieval models have an overlapping goal: increase the capacity of the model to better store, retrieve, and apply knowledge. 
Sparse expert models do this parametrically (i.e. experts contain more learnable parameters), while retrieval based systems embed information that can be dynamically retrieved non-parametrically (i.e. nearest neighbor lookup over a corpus).
Studying the trade-offs and combining both approaches is likely to prove a useful future direction.

\paragraph{Conclusions. }
Sparsity reduces the training and inference costs, resulting in massive models with a better accuracy than their dense counterparts.
But many open questions remain.
For instance, we still poorly understand how the optimal number and size of experts depends on the task (e.g. should one use a few large experts or many small experts for translation?).
As many works have pointed out, achieving strong out-of-domain generalization is less straight-forward and better explanations are needed.
Further, most sparse expert models have relatively low architectural diversity where sparse layers are interspersed at regular intervals.
Future models may benefit from less standardized structure and heterogeneous expert architectures.
Additionally, the appropriate granularity of sparsity still must be determined: most works have focused on experts replacing components, such as feed-forward network layers, but benefits of more fully modular, independent experts were discovered \citep{gururangan2021demix,li2022btm}. 
The field is still uncovering properties of sparse expert models, including much improved calibration \citep{srivastava2022beyond}; others remain unknown including their dynamics under asynchronous training \citep{recht2011hogwild} or their memorization abilities \citep{carlini2020extracting}.
In short, these models pose a myriad of challenging mathematical, engineering, and research problems, but the solutions so far have yielded significant gains and we believe more improvements lie ahead.

\section*{Acknowledgements}
We'd like to thank the BIG-Bench core authors, Mike Lewis, Aidan Clark, Diego de Las Casas, Nan Du, and Carlos Riquelme for permission to reproduce figures and tables here. We would also like to thank Daniel S. Park, Nan Du, Jason Wei, James Lee-Thorp, and Yanqi Zhou for feedback and comments on our drafts.

\clearpage
\bibliographystyle{plainnat}
\bibliography{iclr2020_conference}

\end{document}